\documentclass{sig-alternate}
\usepackage{algorithm}
\usepackage{algpseudocode}

\begin{document}



\title{Deep CTR Prediction in Display Advertising}
%
%
%
%
%
\numberofauthors{5}
\author{
	\eaddfnt{Junxuan Chen\textsuperscript{1,2}\titlenote{The work is done while the author was an intern at Alibaba Group.}, Baigui Sun\textsuperscript{2}, Hao Li\textsuperscript{2}, Hongtao Lu\textsuperscript{1}\titlenote{Corresponding author.}, Xian-Sheng Hua\textsuperscript{2}\titlenote{Corresponding author.}}\\     
	%
	%
	%
	%
	%
	\affaddr{\textsuperscript{1}Department of Computer Science and Engineering, Shanghai Jiao Tong University}\\
	\and
	\affaddr{\textsuperscript{2}Alibaba Group, Hangzhou, China}  \\
	\and
	\affaddr{\{chenjunxuan, htlu\}@sjtu.edu.cn \{baigui.sbg, lihao.lh, xiansheng.hxs\}@alibaba-inc.com}
}
\date{5 July 2016}

\maketitle
\begin{abstract}
	Click through rate (CTR) prediction of image ads is the core task of online display advertising systems,
	and logistic regression (LR) has been frequently applied as the prediction model. However, LR model lacks the ability of extracting complex and intrinsic nonlinear features from handcrafted high-dimensional image features, which limits its effectiveness. To solve this issue, in
	this paper, we introduce a novel deep neural network (DNN) based model that directly predicts the CTR
	of an image ad based on raw image pixels and other basic features in one step. The DNN model employs convolution layers
	to automatically extract representative visual features from images, and nonlinear CTR features are then learned from visual features and other contextual features by using fully-connected layers. Empirical evaluations on a real world dataset with over 50 million records demonstrate the effectiveness and efficiency of this method.
	
\end{abstract}

%
%
\clubpenalty=10000 
\widowpenalty = 10000

%
%

%
%


\keywords{DNN, CNN, Click through rate, Image Ads, Display Advertising}

\section{Introduction}
\begin{figure}
	\centering
	\includegraphics[width=0.45\textwidth]{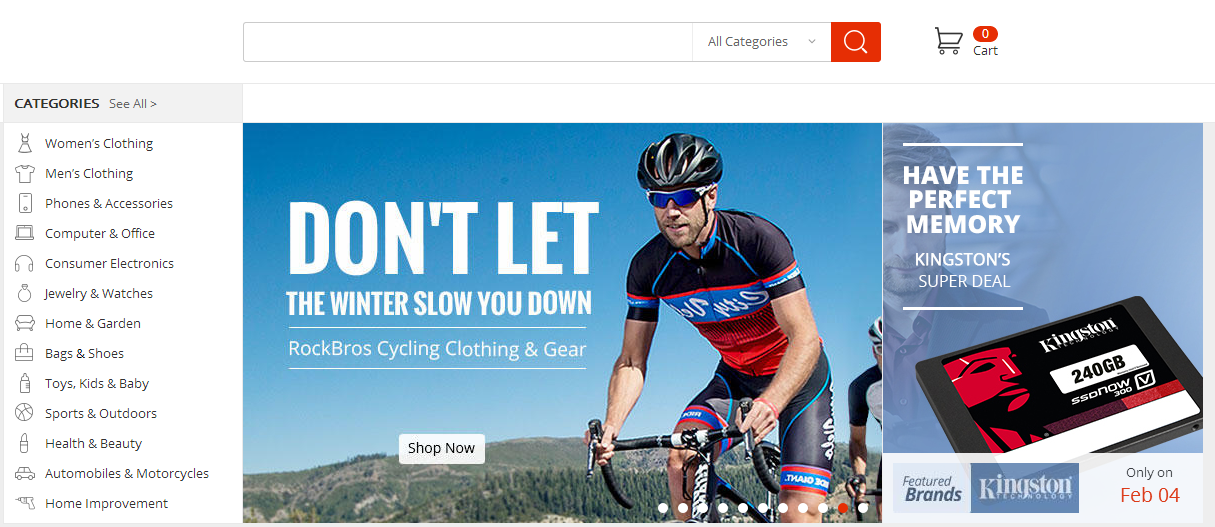}
	\caption{Display ads on an e-commerce web page.}
	\label{fig:ad}
\end{figure}
Online display advertising generates a significant amount of revenue by showing textual or image ads on various web pages \cite{chapelle2014simple}.   The ad publishers like Google and Yahoo sell  ad zones on different web pages to advertisers who want to show their ads to users. And then Publishers get paid by advertisers every time the ad display leads to some desired action  such as clicking or purchasing according to the payment options such as cost-per-click (CPC) or cost-per-conversion (CPA) \cite{mahdian2007pay}. The expected revenue for publishers is the product of the bid price and click-through rate (CTR) or conversion rate (CVR).

Recently, more and more advertisers prefer displaying image ads \cite{mei2011internet} (Figure \ref{fig:ad}) because they are more attractive and comprehensible compared with textual ads.   To maximize the revenue of publishers,  this has led to a huge demand on approaches that are able to choose the most proper image ad to show for a particular user when he or she is visiting a web page so that to maximize the CTR or CVR. 

Therefore, in most online advertising systems, predicting the CTR or CVR is the core task of ads allocation.  In this paper, we focus on CPC and predict the CTR of display ads. Typically an ads system  predicts and ranks the CTR of available ads based on contextual information, and then shows the top $K$ ads to the users. In general, prediction models are learned from past click data based on machine learning techniques \cite{chapelle2014simple,  richardson2007predicting, he2014practical, dave2010learning, zhang2016deep,mcmahan2013ad}.

\begin{figure}
	\centering
	\includegraphics[width=0.45\textwidth]{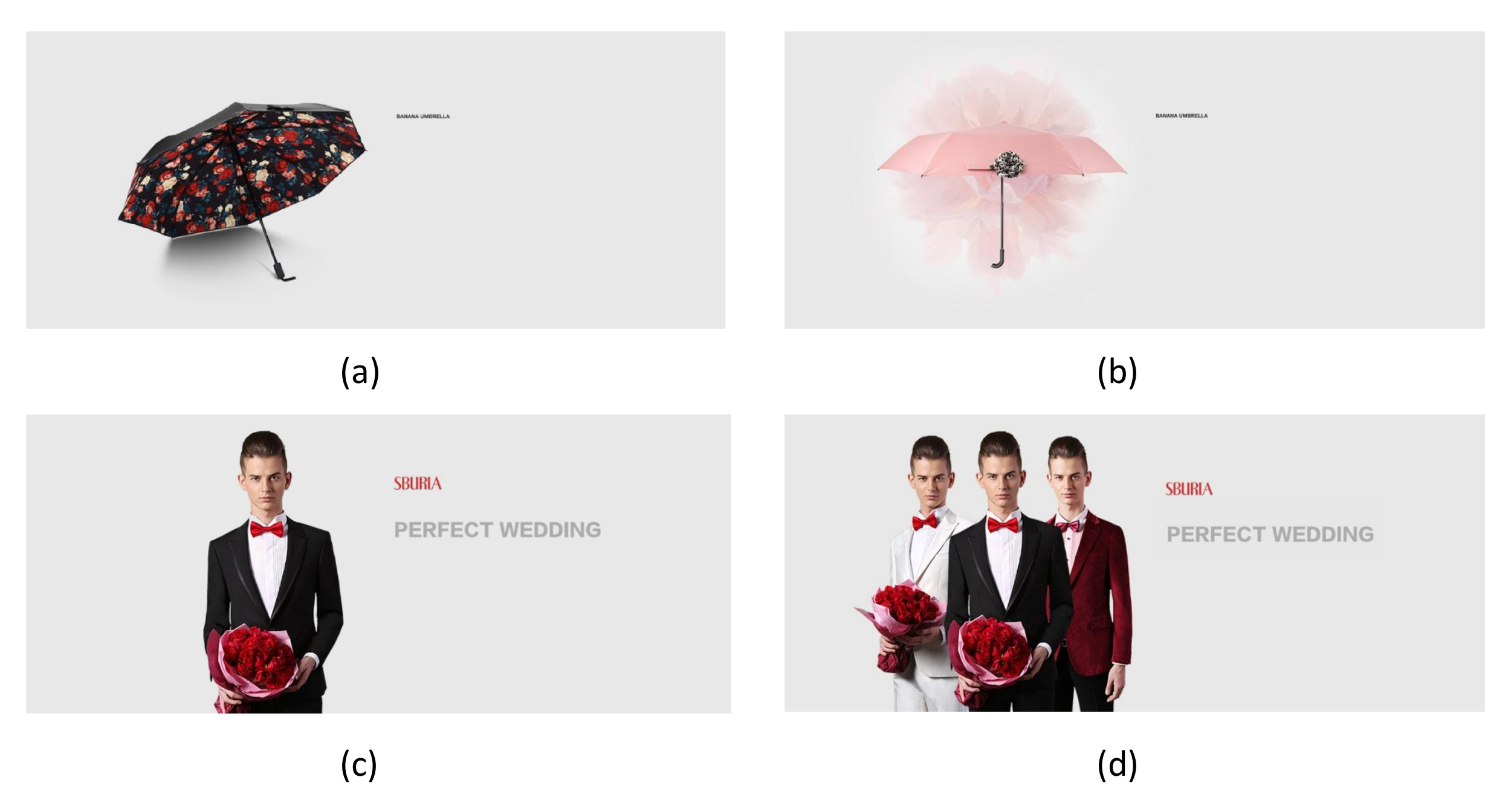}
	\caption{Two groups of image ads in each row. The two ads in each group have completely same ad group id, ad zone and target people. CTRs of image ads (a) and (b) are 1.27\% and 0.83\%. (b) suffers from  low contrast between product and background obviously. CTRs of (c) and (d) are 2.40\% and 2.23\%. We find too many subjects in an men's clothing ad may bring  negative effect. (the number of impressions of each ad is sufficiently high to make the CTR meaningful).}
	\label{fig:compare}
\end{figure}

Features that are used to represent an ad are extremely important in a machine learning model. In recent years, to make the CTR prediction model more accurate, many researchers use millions of features to describe a user's response record (we call it an ad impression). Typically, an image ad impression has basic features and visual features. The basic features are information about users, products and ad positions in a web page, etc. Visual features describe the visual appearance of an image ad at different levels. For example, color and  texture are low level features, while face and other contextual objects are high level features. Low level and high level features may both have the power to influence the CTR of an image ad (Figure \ref{fig:compare}). Traditionally,  researchers lack effective  method to extract  high-level visual features. The importance of visual features is also usually under estimated. However, as we can see from Figure \ref{fig:compare}, ads with same basic features may have largely different CTRs due to different ad images. As a consequence, How to use the visual features in machine learning models effectively becomes an urgent task.

Among different machine learning models that have been applied to predict ads CTR using the above features, Logistic regression (LR) is the mostly well-known and widely-used one due to its simplicity and effectiveness. Also, LR is easy to be parallelized on a distributed computing system thus it is not challenging to make it work on billions of samples \cite{chapelle2014simple}. Being able to handle big data efficiently is necessary for a typical advertising system especially when the prediction model needs to be updated frequently to deal with new ads. However, LR is a linear model which is inferior in extracting complex and effective nonlinear features from handcrafted feature pools. Though one can mitigate this issue by computing the second-order conjunctions of the features, it still can not extract higher-order nonlinear representative features and may cause feature explosion if we continue increasing the conjunction order. 

To address these problems, other methods such as factorization machine \cite{rendle2010factorization}, decision tree \cite{he2014practical}, neural network \cite{zhang2016deep} are widely used. Though these methods can extract non-linear features, they only deal with basic features and handcrafted visual features, which are inferior in describing images. In this paper, we propose a deep neural network (DNN) to directly predict the CTR of an image ad from raw pixels and other basic features. Our DNN model contains convolution layers to extract representative visual features and then fully-connected layers that can learn the complex and effective nonlinear features among basic and visual features. The main contributions of this work can be summarized as follows:
\begin{enumerate}
	\item This paper proposed a DNN model which not only directly takes both high-dimensional sparse feature and image as input, but also can be trained from end to end. To our best knowledge, this is the first DNN based CTR model which can do such things.
	\item Efficient methods are introduced to tackle the challenge of  high-dimensionality and huge data amount in the model training stage. The proposed methods reduce the training time significantly and make it feasible to train on a normal PC with GPUs even with large-scale real world training data.
	\item We conduct extensive experiments on a real-world dataset with more than 50 million user response records to illustrate the improvement provided by our DNN model.  The impacts of several deep learning techniques have also been discussed.  We further visualize the saliency map of image ads to show our model can learn effective visual features.
\end{enumerate}

The paper is organized as follows. Section 2 introduces the related work, followed by an overview of our scheme in Section 3. In Section 4, we describe the proposed DNN model in detail, and we show the challenges in the training stage as well as our solutions  in Section 5. Section 6 presents the experimental results and discussion, and then Section 7 is the conclusion.   
\section{Related Work}
We consider display advertising CTR prediction and deep neural network are two mostly related areas to our work.
\vfill\eject
\subsection{Display Advertising CTR prediction}
Since the display advertising has taken a large share of online advertising market, many works addressing the CTR prediction problem have been published. In \cite{richardson2007predicting,chakrabarti2008contextual}, authors handcraft many features from raw data and use logistic regression (LR) to predict the click-through rate. \cite{chapelle2014simple} also uses LR to deal with the CTR problem and scales it to
billions of samples and millions of parameters on a distributed learning system.  In \cite{oentaryo2014predicting},  a Hierarchical Importance-aware Factorization Machine (FM)  \cite{rendle2012factorization} is introduced, which provides a generic latent factor framework that incorporates importance weights and hierarchical learning. In \cite{dave2010learning}, boosted decision trees have been used to build a prediction model. In \cite{he2014practical},  a model which combines decision
trees with logistic regression has been proposed, and outperforms either of the above two models. \cite{zhang2016deep} combines deep neural networks with FM and also brings  an improvement. All of these methods are very effective when deal with ads without images. However, when it comes to the image ads, they can only use pre-extracted image features, which is less flexible to take account of the unique properties of different datasets.

Therefore, the image features in display advertisement have received more and more attention. In \cite{azimi2012impact,cheng2012multimedia}, the impact of  visual appearance  on user's response in online display advertising is considered for the first time. They extract over 30 handcrafted features from ad images and build a CTR prediction model using image and basic  features. The experiment result shows that their method achieves better performance than models without visual features.   \cite{Mo:2015:IFL:2832747.2832769} is the most related work  in literature with us, in which a decapitated convolutional neural network (CNN) is used to extract image features from ads. However, there are two important differences between their method and ours.   First, they do not consider basic features when extracting image features using CNN. Second,  when predicting the CTR they  use logistic regression which lacks the ability in exploring the complex relations between image and basic features. Most of the information in their image features is redundant  such as product category which is included in basic features. As a result, their model only achieves limited improvements when combining both kinds of features. Worse still, when the dataset contains too many categories of products, it can hardly converge when training. Our model uses an end to end model to predict the CTR of image ads using basic features and raw images in one step, in which image features can be seen as  supplementary to the basic features.
\subsection{Deep Neural Network} 
In recent years, deep neural network has achieved big breakthroughs in many fields. In computer vision field, convolutional neural network (CNN) \cite{NIPS2012_4824} is one of the most efficient tools to extract effective image features from raw image pixels.  In speech recognition, deep belief network (DBN) \cite{hinton2012deep} is used and much better performance is obtained comparing with Gaussian mixture models. Comparing with traditional models that have shallow structure, deep learning can model the underlying patterns  from massive and complex data. With such learning ability, deep learning can be used as a good feature extractor and applied into many other applications \cite{ren2015faster,simonyan2014two}.

In CTR prediction field, besides \cite{zhang2016deep} that is mentioned in Section 2.1, DNN has also been used in some public CTR prediction  competitions\footnote{https://www.kaggle.com/c/avazu-ctr-prediction}\footnote{https://www.kaggle.com/c/criteo-display-ad-challenge} recently. In these two competitions, only basic features are available for participants. An ensemble of four-layer DNNs which use fully-connected layers and different kinds of non-linear activations  achieves better or comparable performance than LR with feature conjunction, factorization machines, decision trees, etc. Comparing with this method, our model can extract more powerful features by taking consideration  of the visual features in image ads. 
\section{Method Overview}\label{Overview}
\begin{figure}
	\centering
	\includegraphics[width=0.45\textwidth]{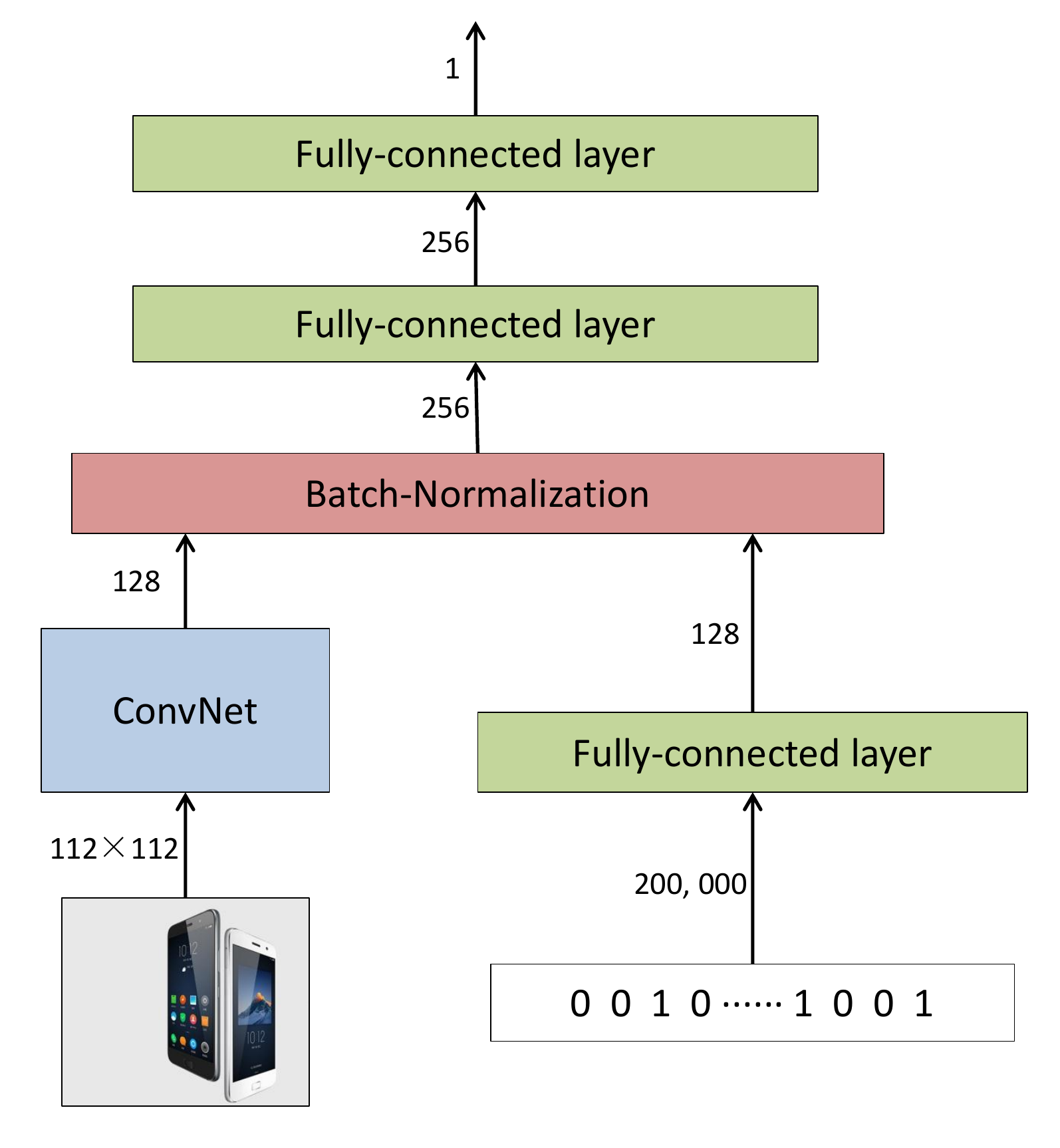}
	\caption{The overall architecture of the network. The output of each fully-connected layer is then pass through a ReLU nonlinear activation function.}
	\label{fig:net}
\end{figure}
As aforementioned,
in this paper, each record of user's behavior on an ad is called an impression. denoted by $x$. Each impression has an image $u$ with a resolution of around $120 \times 200$. Besides the image, the basic feature vector is denoted by $v \in \mathbf{R}^{d}$ such as the  user's gender, product's category, ad position in the web page, and usually $d$ can be very large, say, from a few thousand to many million.
Our goal is to predict the probability that a user clicks on an image ad given these features. We will still use logistic regression to map our predicted CTR value $\hat{y}$ to 0 to 1, thus the CTR prediction problem can be written as:
\begin{gather}
	\hat{y} = \frac{1}{1+e^{-z}} \\
	z = f(x)
\end{gather}
where $f(.)$ is what we are going to learn from training data, that is, the embedding function that maps an impression to a real value $z$.  Suppose we have $N$ impressions $\mathbf{X}=[x_{1}, x_{2}...x_{N}]$ and each with a label $y_{i} \in \{0,1\}$ depends on the user's feedback, 0 means \emph{not clicked} while 1 means \emph{clicked}. Then the learning problem is defined as minimizing a Logarithmic Loss (Logloss):
\begin{equation}
	L(\mathbf{W}) = -\frac{1}{N} \sum_{i} (y_{i} \log \hat{y}_{i} + (1-y_{i}) \log (1-\hat{y}_{i}) )  \\ + \lambda || \mathbf{W}||^{2}
\end{equation}
where $\mathbf{W}$ is the parameters of the embedding function $f(.) $ and $\lambda$ is a regularization parameter that controls the model complexity.

In this model, what we need to learn is the embedding function $f(.)$. Conventional methods extract handcrafted visual features from raw image $u$ and concatenate them with basic features $v$, then learn linear or nonlinear transformations to obtain the embedding function. In this paper we learn this function directly from raw pixels of an image ad and the basic features using one integrated deep neural network.  
\section{Network Architecture}
\begin{figure*}
	\centering
	\includegraphics[width=0.8\textwidth]{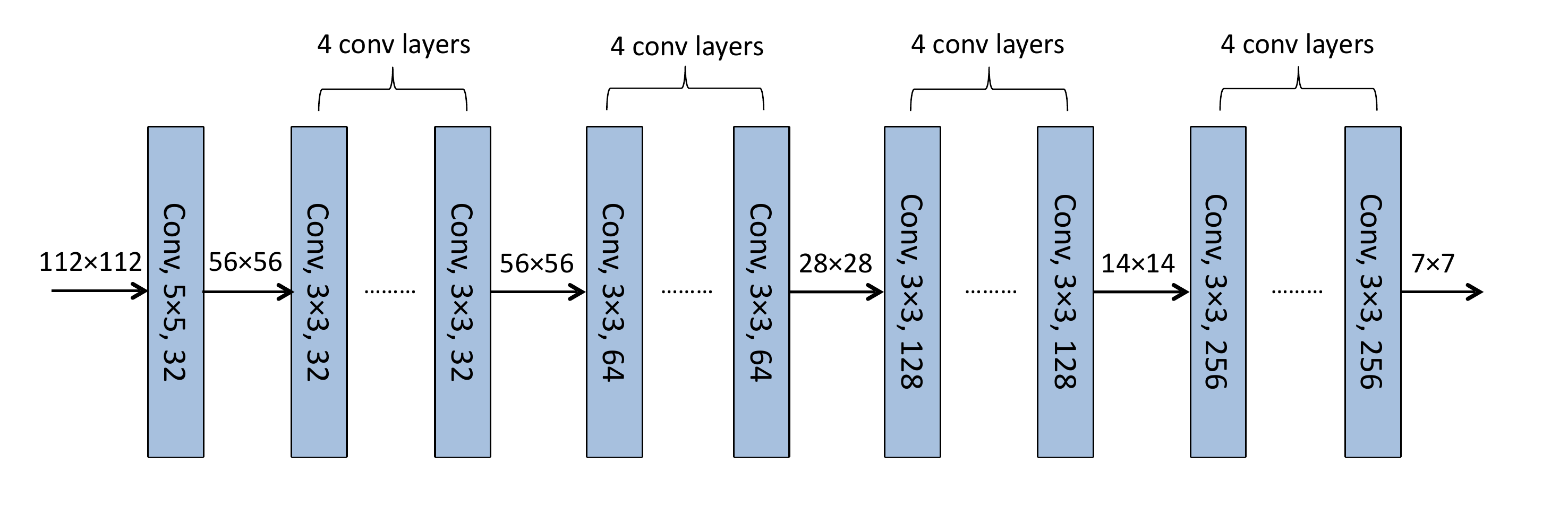}
	\caption{The architecture of the 17-layer \emph{Convnet} in our model.}
	\label{fig:convnet}
\end{figure*}
Considering basic features and raw images come from two different domains, we cannot simply concatenate them together directly in the network. Training two separate networks is also inferior since it cannot take into account the correlations between the two features. As a result, our network adopts two different sub-networks to deal with basic features and raw images, respectively, and then uses multiple fully-connected layers to capture their correlations. 

As illustrated in Figure \ref{fig:net},  a  deep neural network called DeepCTR is designed which contains three parts. 
One part, \emph{Convnet}, takes raw image $u$ as input and follows with a convolutional network. 
The output of the \emph{Convnet} is a feature vector of the raw image. 
The second part which is called \emph{Basicnet}, takes basic features $v$ as input and applies a fully-connected layer to reduce the dimensionality. 
Subsequently, outputs of \emph{Convnet} and \emph{Basicnet} are concatenated into one vector and fed to two fully-connected layers. 
The output of the last fully-connected layer is a real value $z$.
This part is called \emph{Combnet}. On the top of the whole network, Logloss is computed as described in Section \ref{Overview}.

The design of \emph{Convnet} is inspired by the network in \cite{he2015deep,simonyan2014very}, as shown in Figure \ref{fig:convnet}. The network consists of 17 convolution layers. The first convolution layer uses $5 \times 5$ convolution kernels. Following first layer, there are four groups and each has four layers with $3 \times 3$ kernels.  We do not build a very deep network such as more than 50 layers in consideration of the trade off between performance and training time. We pre-train the \emph{Convnet} on the images in training dataset with category labels. We use two fully-connected layers with 1024 hidden nodes (we call them \emph{fc18} and \emph{fc19}), a fully-connected layer with 96-way outputs (we call it \emph{fc20}) and a softmax after the \emph{Convnet} in pre-training period.  Since our unique images set is smaller (to be detailed in Section 6) than ImageNet \cite{deng2009imagenet}, we use half the number of outputs in each group comparing with \cite{he2015deep}. After pre-training,  a 128-way fully-connected layer is connected behind the last convolution layer.  Then we train the whole DeepCTR using Logloss from end to end.
\section{Speed Up Training}
An online advertising system has a large number of new user response records everyday. It is necessary for ad systems  to update as frequently as possible to adapt  new tendency. An LR model with distributed system requires several hours to train with billions of samples, which makes it popular in industry.

Typically a deep neural network has millions of parameters which makes it impossible to train quickly. With the development of GPUs, one can train a deep CNN with 1 million training images in two days on a single machine. However, it is not time feasible for our network since we have more than 50 million samples. Moreover, the dimensionality of basic features is nearly 200,000 which leads to much more parameters in our network than a normal deep neural network. Directly training our network on a single machine may take hundreds of days to converge according to a rough estimation. Even using multi-machine can hardly resolve the training problem. We must largely speed up the training if we want to deploy our DeepCTR on a real online system. 

To make it feasible to train our model with less than one day, we adopt two techniques: using sparse fully-connected layer and a new data sampling scheme. The use of these two techniques makes the training time of our DeepCTR suitable for a real online system.
\subsection{Sparse Fully-Connected Layer}
In CTR prediction, the basic feature of an ad impression includes user information like gender, age, purchasing power, and ad information like ad ID, ad category, ad zone, etc. This information is usually encoded by one-hot encoding or feature hashing \cite{weinberger2009feature} which makes the feature dimension very large. For example, it is nearly 200,000 in our dataset. Consequently, in \emph{Basicnet}, the first fully-connected layer using the basic feature as input has around 60 million parameters, which is similar to the number of all the parameters in  AlexNet \cite{NIPS2012_4824}. However, the basic feature is extremely sparse due to the one-hot encoding. Using sparse matrix in first fully-connected layer can largely reduce the computing complexity and GPU memory usage.

In our model, we use compressed sparse row (CSR) format to represent a batch of basic features $V$. When computing network forward 
\begin{equation}
	Y_{fc1} = VW,
\end{equation}
sparse matrix operations  can be used in the first fully-connected layer. When backward pass,  we only need to update the weights that link to a small number of nonzero dimensions according to the gradient
\begin{equation}
	\nabla(W) = V.
\end{equation}
Both of the forward pass and backward pass only need a time complexity of $O(nd')$ where $d'$ is the number of  non-zero elements in basic features and $d' \ll d$. An experiment result that compares the usages of time and GPU memory with/out sparse fully-connected layer can be found in Section 6.2.
\subsection{Data Sampling}
Another crucial issue in  training is that \emph{Convnet} limits the batch-size of Stochastic Gradient Descent (SGD). To train a robust CTR prediction model, we usually need  millions of ad samples. However, the \emph{Convnet} requires lots of GPU memory, which makes our batch-size very small, say, a few hundred. For a smaller batch-size, the parallel computing of GPU can not maximize the effect in the multiplication of large matrix. And the number of iterations of each epoch will be very large. We need much more time to run over an epoch of iterations. Though the sparse fully-connected layer can largely reduce the forward-backward time in \emph{Basicnet}, training the whole net on such a large dataset still requires infeasible time. Also, the gradient of each iteration is unstable in the case of smaller batch-size, which makes the convergence harder. In CTR prediction, this problem is even more serious because the training data is full of noise.

\begin{figure}
	\centering
	\includegraphics[width=0.4\textwidth]{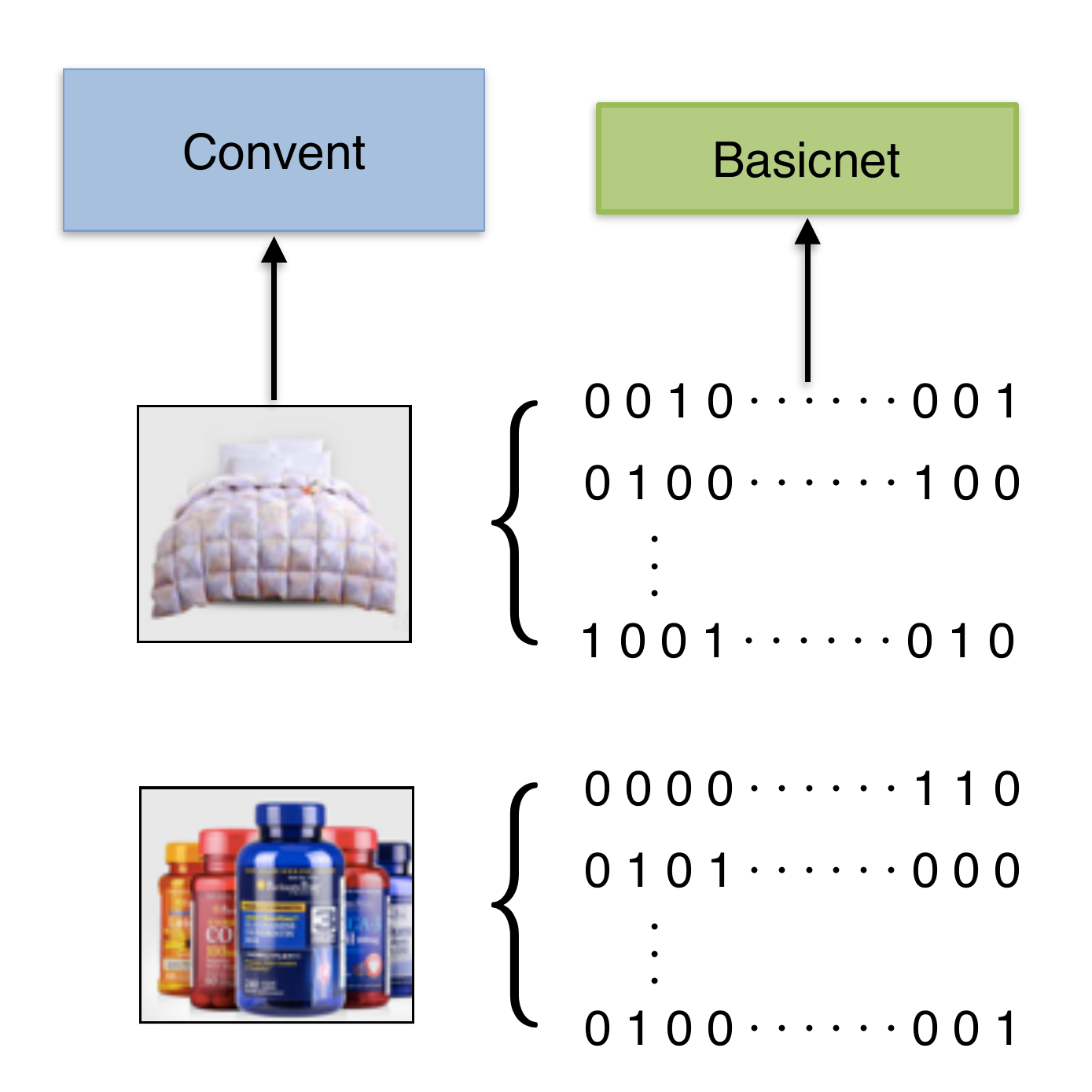}
	\caption{Originally, the image and the basic feature vector are one to one correspondence. In our data sampling method, we group basic features of an image together, so that we can deal with much more basic features per batch. }
	\label{fig:sampling}
\end{figure}

In this paper, we propose a simple but effective training method based on an intrinsic property of the image ads click records, that is, many impressions share a same image ad. Though the total size of the dataset is very large, the number of unique images is relatively smaller. Since a good many of basic features can be processed quickly by sparse fully-connected layer, we can set a larger batch-size for \emph{Basicnet} and a smaller one for \emph{Convnet}. In this paper we employ a data sampling method that groups basic features of a same image ad together to achieve that (Figure \ref{fig:sampling}), which is detailed as follows.

Suppose the unique  images set in our dataset is $\mathbf{U}$, the set of impressions related to an image $u$ are $\mathbf{X}_{u}$ and basic features are $\mathbf{V}_{u}$. At each iteration, suppose the training batch size is $n$, we sample $n$ different images $U$ from $\mathbf{U}$. Together with each image $u \in U$, we sample $k$ basic features $V_{u}$ from $\mathbf{V}_{u}$ with replacement. Thus we have $n$ images and $kn$ basic features in each batch.   After \emph{Convnet}, we have $n$ image features. For each feature vector $conv_{u}$ we copy it $k$ times to have $C_{u}$ and send them forward to \emph{Combnet} along with $V_{u}$. In backward time, the gradient of each image feature vector can be computed as:
\begin{equation}\label{eq:gradient}
	\nabla(conv_{u}) = \frac{1}{k} \sum_{c \in C_{u}} \nabla(c)  
\end{equation}
The training method is summarized in Alg. \ref{alg:training} and Alg. \ref{alg:fb}. In fact, this strategy makes us able to deal with $kn$ samples in a batch. Since the sparse fully-connected layer requires very small GPU memory, we can set $k$ a very big value according to the overall average number of basic feature vectors of image ads. This strategy reduces the number of iterations of an epoch to several thousand and largely  speeds up training. A larger batch-size also makes the gradient of each batch much more stable which leads to the model easy to converge. We also conduct an experiment to evaluate whether this sampling method influences  the performance of DeepCTR comparing a throughly shuffle strategy in Section  6.2.
\begin{algorithm}[tb]
	\caption{Training a DeepCTR network}
	\label{alg:training}
	\begin{algorithmic}[1]
		\renewcommand{\algorithmicrequire}{\textbf{Input:}}
		\renewcommand{\algorithmicensure}{\textbf{Output:}}
		\Require: Network $Net$ with parameter $\mathbf{W}$, unique images set $\mathbf{U}$, basic features set $\mathbf{V}$, labels $\mathbf{Y}$, batch size $n$, basic feature sample number $k$.  
		\Ensure: Network for CTR prediction, $Net$
		\State Initialize $Net$.
		\State Compute the sample probability $p(u)$ of each image $u$,
		\begin{equation}
			p(u) = \frac{\# \mathbf{V}_{u}}{\sum_{u' \in \mathbf{U}} \# \mathbf{V}_{u'}}
		\end{equation}
		\Repeat
		\State Sample $n$ images $U$ according to $p(u)$. 
		\State For each $u$ in $U$, sample $k$ basic features $V_{u}$ from $\mathbf{V}_{u}$ with labels $Y_{u}$ uniformly with replacement. 
		\State $forward\_backward(Net, U, V, Y)$.
		\Until{$Net$ converges}
	\end{algorithmic}
\end{algorithm}  

\begin{algorithm}[tb]
	\caption{$forward\_backward$}
	\label{alg:fb}
	\begin{algorithmic}[1]
		\renewcommand{\algorithmicrequire}{\textbf{Input:}}
		\renewcommand{\algorithmicensure}{\textbf{Output:}}
		\Require: Network $Net$ with parameters $\mathbf{W}$ which contains a \emph{Convnet}, \emph{Basicnet} and \emph{Combnet}, image samples $U$, basic features $V$, labels $Y$, basic feature sample number $k$.

		\State Compute the feature vector $conv_{u}$ of each image $u$:
		$
		conv = net\_foward(Convnet, U)
		$
		\State Copy each feature vector $k$ times so we have $C$.
		
		\State  $loss = net\_forward(Basicnet \text{ and } Combnet, V, C)$.
		\State  $\nabla(C) = net\_backward(Combnet \text{ and } Basicnet, loss)$.
		\State Compute $\nabla(conv_{u})$ of each image $u$ according to Eq. \ref{eq:gradient}.
		\State $net\_backward(Convnet, \nabla(conv))$.
		\State Update network $Net$.
	\end{algorithmic}
\end{algorithm}  
\section{Experiment}
In this section, a series of experiments are conducted to verify the superiority of our DeepCTR model.
\subsection{Experimental Setting}
\subsubsection{Dataset}
The experiment data comes from a commercial advertising platform in an  arbitrary  week of year 2015. We use the data from first six days as our training data and the data from last day (which is a Friday) as testing data.  As described in Section \ref{Overview}, each impression consists of an ad $x$ and a label $y$.  An impression has an image $u$ (Figure \ref{fig:compare}) and a basic feature vector $v$.  The size of training data is 50 million while testing set is 9 million. The ratio of positive samples and negative samples is around 1:30.   We do not perform any sub-sampling of negative events on the dataset. We have 101,232 unique images in training data and 17,728 unique images in testing data. 3,090 images in testing set are never shown in training set. Though the image data of training set and test data are highly overlapped, they follow the distribution of the real-world data. To make our experiment more convincing, we also conduct a experiment on a sub test set that only contains new images data that never been used in training.  The basic feature $v$ is one-hot encoded and has a dimensionality of 153,231. Following information is consisted by basic features:
\begin{enumerate}
	\item Ad zone. The display zone of an ad on the web page. We have around 700 different ad zones in web pages.
	\item Ad group. The ad group is a small set of ads. The ads in an ad group share almost same  products but different ad images (in Figure \ref{fig:compare}, (a) and (b) belong to an ad group while (c) and (d) belong to another group). We have over 150,000 different ad groups in our dataset. Each ad group consists less than 20 different   ads.
	\item Ad target. The groups of target people of the ad.  We have 10 target groups in total.
	\item Ad category. The category of the product in ads. We have 96 different categories, like clothing, food, household appliances. 
	\item User. The user information includes user's gender, age, purchasing power, etc. 
\end{enumerate}
Besides above basic features, we do not use any handcrafted conjunction features. We hope that  our model can learn  effective non-linear features automatically from feature pools.

\subsubsection{Baselines}
We use LR only with basic features as our first baseline. We call this method \emph{lr basic} in following experiments. To verify that our DNN model has the ability of extracting effective high-order features,  a Factorization Machine implemented by LibFM \cite{rendle2012factorization}  only with basic features is our second baseline.  We call it \emph{FM basic}. We use 8 factors for 2-way interactions and MCMC for parameter learning in \emph{FM basic}. Then we  evaluate a two hidden layers DNN model only using basic features. The numbers of outputs of two hidden layers  are 128 and 256 respectively. The model can be seen as our DeepCTR net without the \emph{Convnet} part. This method is called \emph{dnn basic}. We further replace the \emph{Convnet} in our DeepCTR net with pre-extracted features, SIFT \cite{lowe1999object} with bag of words and the outputs of different layers of the pre-trained  \emph{Convnet}. We call these two methods \emph{dnn sift} and \emph{dnn layername} (for example \emph{dnn conv17}). 
\subsubsection{Evaluation Metric}
We use two popular metrics to evaluate the experiment result, Logloss and the area under receiver operator curve (AUC).  Logloss can quantify the accuracy of the predicted click probability. AUC measures the ranking quality of the prediction. Our dataset comes from a real commercial platform, so both of these metrics use relative numbers comparing with \emph{lr basic}. 

Since the AUC value is always larger than 0.5, we remove this constant part (0.5) from the AUC value and then compute the relative numbers as in \cite{yan2014coupled}:
\begin{equation}
	\text{relative AUC} = (\frac{AUC(method) - 0.5}{AUC(\emph{lr basic})-0.5}-1) \times 100\% 
\end{equation}  
\subsubsection{Network Configuration}
In our \emph{Convnet},  a $112 \times 112$ random crop  and horizontal mirror for the input image are used for data augmentation. Each group has four convolution layers followed by a batch normalization \cite{ioffe2015batch} and a ReLU \cite{nair2010rectified} activation. The stride of the first convolution layer is 2 if the output size of a group halves.  We initialize the layer weights as in  \cite{he2015delving}. When pre-training the \emph{Convnet} on our image dataset with category labels,  we use SGD with a mini-batch size of 128. The learning rate starts from 0.01 and is divided by 10 when test loss plateaus. The pre-trained model converges after around 120 epochs. The weight decay of the net is set as 0.0001 and momentum is 0.9. 

After pre-training  \emph{Convnet}, we train our DeepCTR model from end to end. Other parts of our net use the same  initialization method as  \emph{Convnet}. We choose the size of mini-batch $n$ as 20, and $k = 500$. That is to say, we deal with 10,000 impressions per batch.  We start with the learning rate 0.1, and divided it by 10 after $6\times10^{4}$, $1\times10^{5}$ and $1.4\times10^{5}$ iterations. The \emph{Convnet} uses a smaller initial learning rate 0.001 in case of destroying the pre-trained model. The weight decay of the whole net is set as $5\times 10^{-5}$. The \emph{dnn basic}, \emph{dnn sift} and \emph{dnn layername} use the same learning strategy. 

We implement our deep network on C++ Caffe toolbox \cite{jia2014caffe} with some modifications like sparse fully-connected layer.
\begin{table*}
	\centering
	\caption{relative AUC and Logloss. All the numbers are   best resuts achieved in three repeated experiments. We omit \emph{dnn} of methods using deep neural network with pre-extracted features.}
	\label{table:AUC}
	\begin{tabular}{|c|c c c c|c c c c c|c c|} \hline
		method  & lr basic &FM basic &  basic &sift &conv13 &conv17 &fc18 &fc19 &fc20 &DeepCTR &3 DeepCTRs \\  \hline
		AUC(\%) & -        &  1.47 & 1.41  &1.69 &  3.92 &4.13   &4.10 &3.48 &3.04 &5.07    &\textbf{5.92} \\  \hline
		Logloss(\%)&-      &  -0.40 & -0.39 &-0.45& -0.79 & -0.86 &-0.86&-0.74&-0.69&-1.11   & \textbf{-1.30}\\ \hline
	\end{tabular}
\end{table*}

\subsection{Results and Discussion}
\begin{figure}
	\centering
	\includegraphics[width=0.45\textwidth]{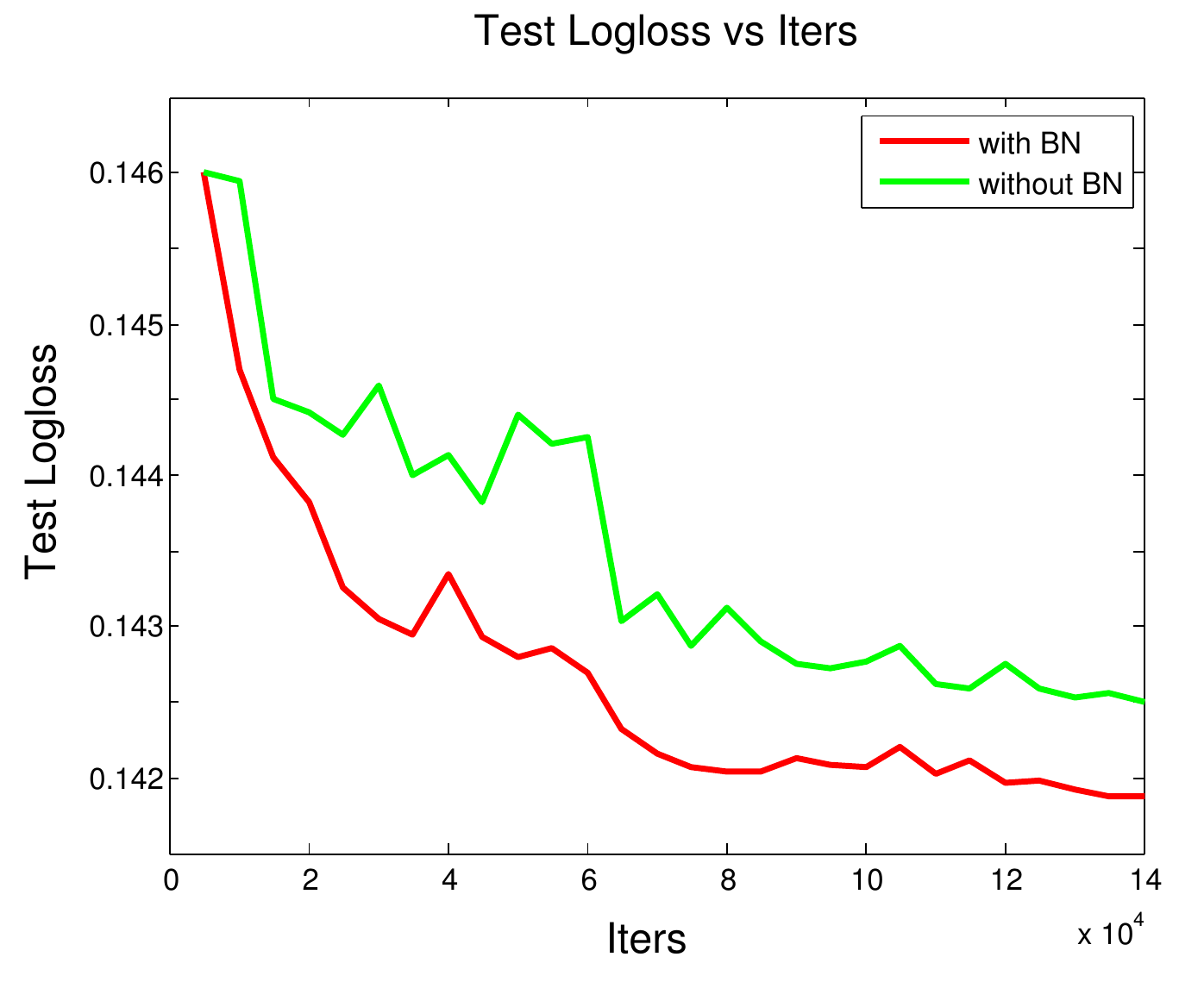}
	\caption{Test Logloss of the DeepCTR net with/without batch normalization in \emph{Combnet}. }
	\label{fig:bn}
\end{figure} 
In this section we compare the results of various methods and the effects of some network structures. First we compare the results of  models with deep features in different levels. We plot the two metrics of \emph{dnn conv13}, \emph{dnn conv17}, \emph{dnn fc18}, \emph{dnn fc19}, and \emph{dnn fc20} in the middle of Table \ref{table:AUC}. From the results, we find that \emph{dnn conv17} and \emph{dnn fc18} achieve best performance. Image features in these layers are of relatively high level but not highly group invariant \cite{zeiler2014visualizing}. Comparing with following fully-connected layers, they have more discriminations in same category. Comparing with previous layers, they contain  features in a sufficiently high-level  which are superior in describing the objects in images. Consequently, we connect \emph{conv17} layer in our DeepCTR model. We do not choose \emph{fc18} because it needs higher computations. We have also tried to compare our DeepCTR with the approach in \cite{Mo:2015:IFL:2832747.2832769}. However the model in \cite{Mo:2015:IFL:2832747.2832769} does not converge on our dataset. We think the reason is that our dataset consists of too many categories of products while in their datasets only 5 different categories are available.
\begin{figure}
	\centering
	\includegraphics[width=0.45\textwidth]{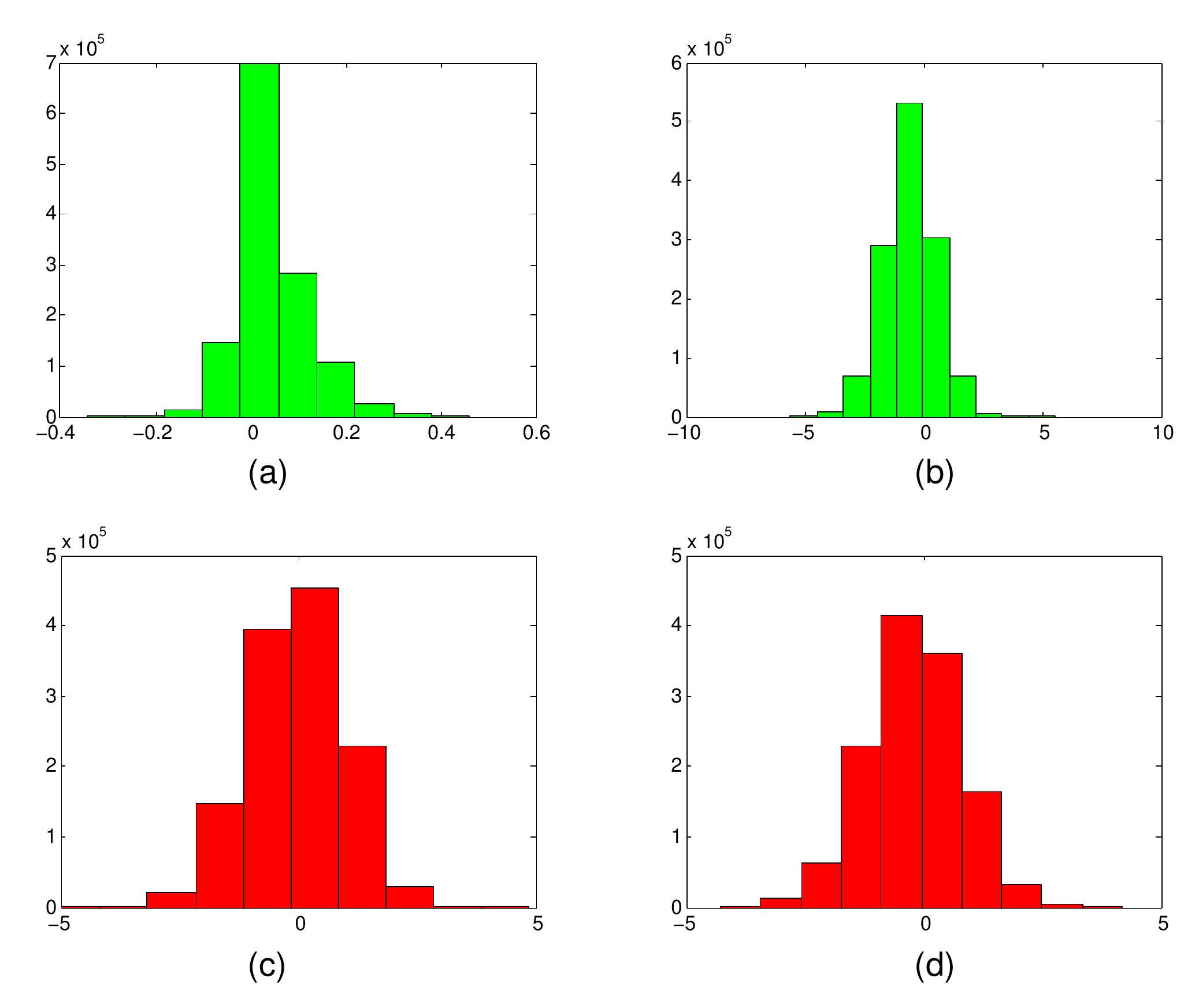}
	\caption{(a) and (b) are the histograms of outputs of \emph{Basicnet} and \emph{Convnet} without batch normalization while (c) and (d) with batch normalization. }
	\label{fig:bn_comb}
\end{figure}

Comparison between other baselines is shown in Table \ref{table:AUC} too. From the result, it can be seen that a deep neural network and image features can both improve the CTR prediction accuracy.  \emph{FM basic} and \emph{dnn basic} achieve almost same improvements comparing with \emph{lr basic}, which  indicates that these two models both have strong  power in extracting effective non-linear basic features.   For the image part, comparing with handcrafted features, like SIFT, deep features have stronger power in describing the image, which leads to a significant improvement in the prediction accuracy. Our DeepCTR model goes one step further by using an end to end learning scheme. Ensemble of multiple deep networks usually brings better performance, so we train 3 DeepCTR models and average their predictions, and it gives the best AUC and Logloss. Compared with \emph{lr basic}, the AUC increase will bring us 1$\sim$2 percent CTR increase in the advertising system (according to online experiments), which will lead to over 1 million earnings growth per day for an 100 million ads business.  

\begin{table}
	\centering
	\caption{relative AUC and Logloss of the sub test set that only contains images never shown in the training set.}
	\label{table:subset}
	\begin{tabular}{|c|c|c|} \hline
		& AUC (\%) & Logloss (\%)\\ \hline
		lr basic & - & - \\ \hline
		dnn basic & 1.07 & -0.21 \\ \hline
		dnn sift & 2.14 & -0.48  \\ \hline
		DeepCTR & \textbf{5.54} & \textbf{-0.85} \\ \hline
	\end{tabular}
\end{table}
To make our results more convincing, we also conduct an experiment on the sub test set that only contains 3,090 images that are never shown in training set. The relative AUC and Logloss of three representative methods \emph{dnn basic}, \emph{dnn sift} and a single DeepCTR are in Table \ref{table:subset}. Clearly, our DeepCTR wins by a large margin consistently. We also notice that while the AUC of \emph{dnn basic} decreases, \emph{dnn sift} and our DeepCTR have an even higher relative AUC than the result on the full test set. 
Ad images in 3K sub test set are all new ads added into the ad system and the ad groups have not appeared in the training set. This lead to the prediction worse in 3K sub test set because the ad group feature does not exist in the training set. However, though we lack some basic features, visual features bring much more supplementary information. This is the reason that \emph{dnn sift} and DeepCTR have more improvements over the baseline methods (which only have basic features) in 3K sub test set comparing with the full test set.
This experiment shows that visual features can be used to identify ads with similar characteristics and thus to predict the CTR of new image ads  more accurately.   It also verifies that our model indeed has strong generalization ability but not  memories the image id rigidly.

For different real-world problems, different techniques may be needed due to the intrinsic characteristics of the problem, network design and data distribution. Therefore, solutions based on deep learning typically will compare and analyze the impact and effectiveness of those techniques to find the best practices for a particular problem. Therefore,
we further  explore the influence of different deep learning techniques in our DeepCTR model empirically. 

First we find that the batch normalization in the \emph{Combnet} can speed up training and largely improve performance (Figure \ref{fig:bn}). To investigate the reason, we show the histogram (Figure \ref{fig:bn_comb}) of the outputs of \emph{Convnet} and $BasicNet$. We can see from the histogram  that two outputs have significant difference in scale and variance. Simply concatenating  these two different kinds of data stream  makes the following fully-connected layer hard to converge. 
\begin{figure}
	\centering
	\includegraphics[width=0.45\textwidth]{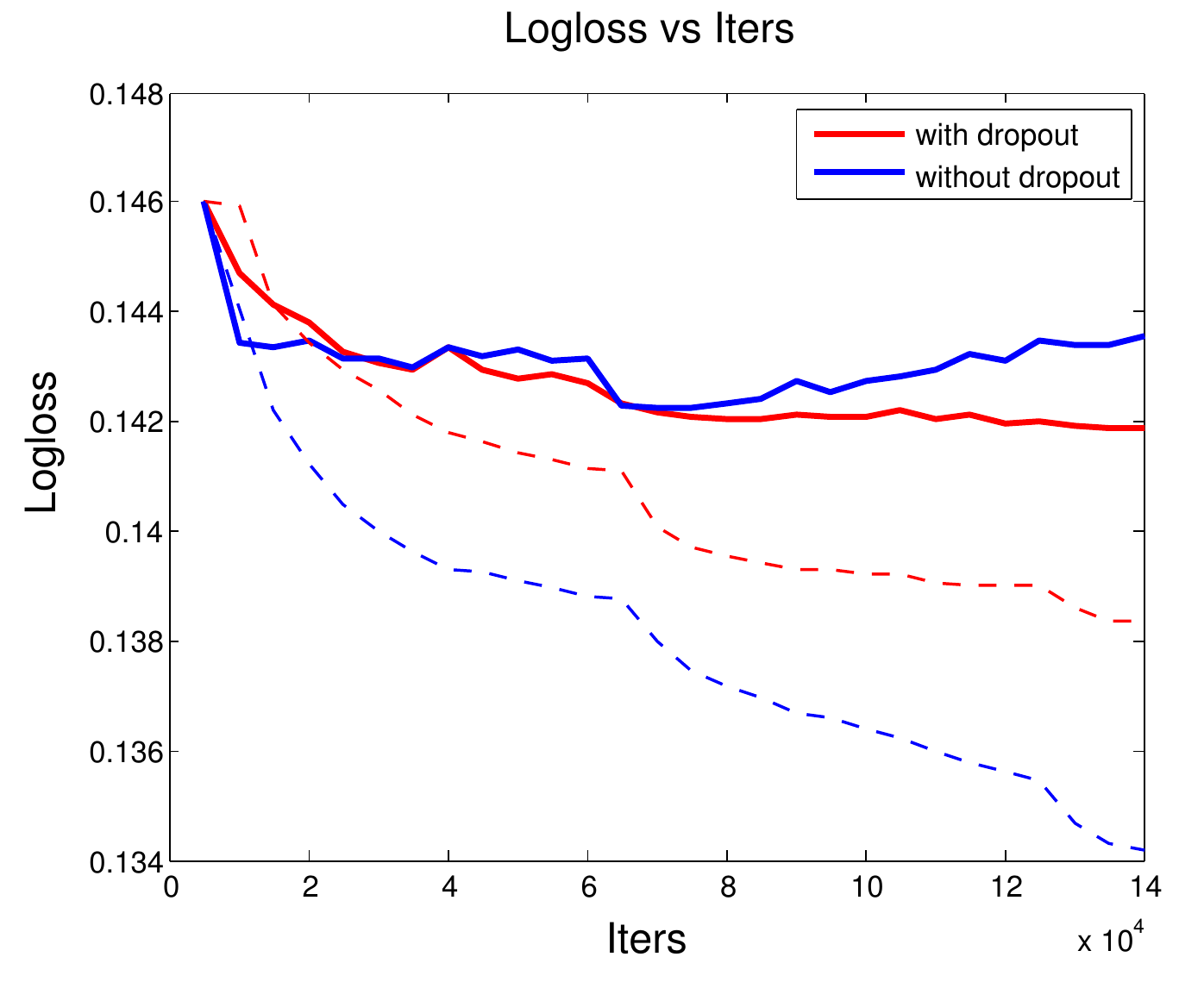}
	\caption{Logloss of the DeepCTR net with/without dropout in \emph{Combnet}. Dashed lines denote training loss, and bold lines denote test loss.}
	\label{fig:dropout}
\end{figure}

Dropout \cite{srivastava2014dropout} is an efficient way to prevent over-fitting problem in deep neural network. Most deep convolution networks remove the dropout because batch normalization can regularize the models \cite{he2015deep, ioffe2015batch}. However, in our DeepCTR model, we find it still suffers from over-fitting without dropout. We compare the loss curves of the model with/without dropout in the last two fully-connected layers. We can see that the model with dropout achieves lower testing Logloss, though we need more time to reach the lowest test loss.

We also evaluate the performance of the sparse fully-connected layer and our data sampling method. We plot computing time and memory overhead (Table \ref{table:time&memory}) of the sparse fully-connected layer comparing with dense layer. Loss curves of training and testing are exactly the same since sparse fully-connected layer does not change any computing results in the net, so we do not plot them. From this table we can find dense layer  requires much more computing time and memory than sparse one. Using sparse layer allows a lager batch size when training, which speeds up the training and makes the net much easier to converge. 

\begin{table}
	\centering
	\caption{forward-backward time and GPU memory overhead of first fully-connected layer with a batch size of 1,000.}
	\label{table:time&memory}
	\begin{tabular}{|c|c|c|} \hline
		& time (ms) & memory (MB)\\ \hline
		sparse layer & 6.67 &397 \\ \hline
		dense layer & 189.56 &4667 \\ \hline
	\end{tabular}
\end{table}

Finally, we investigate whether the performance of our model descends using our data sampling method comparing a throughly shuffle. We only evaluate the sampling method on  \emph{dnn conv17} model, that is, we conduct experiments on a model where the \emph{Convnet} is frozen. Ideally, we should use an unfrozen \emph{Convnet} without data sampling as the contrast experiment. However, as mentioned in Section 5.2, training an unfrozen \emph{Convnet} limits our batch-size less than 200  because the \emph{Convnet} needs much more GPU memory, while a model with frozen \emph{Convnet} can deal with more than 10000 samples in a batch. It will takes too much time to training our model on such a small batch-size. Also, the main difference between with/out sampling is whether the samples were thoroughly shuffled, while freezing the \emph{Convnet} or not does not influence  the order of samples. Therefore, we believe that our  DeepCTR model performs similarly with \emph{dnn conv17} model. From Table \ref{table:datasampling} we can see the performance of the model is not influenced by the data sampling method. At the same time, our method  costs far less training time comparing with the approach without data sampling. Using this data sampling method, training our DeepCTR model from end to end only takes around 12 hours to converge on a NVIDIA TESLA k20m GPU with 5 GB memory, which is acceptable for an online advertising system requiring daily update. 
\begin{table}
	\centering
	\caption{AUC and Logloss of \emph{dnn conv17} model with our data sampling and a throughly shuffle.}
	\label{table:datasampling}
	\begin{tabular}{|c|c|c|} \hline
		& AUC(\%) & Logloss(\%)\\ \hline
		data sampling & 4.13 & -0.86\\ \hline
		throughly shuffle & 4.12 & -0.86\\ \hline
	\end{tabular}
\end{table}

\begin{figure}
	\centering
	\includegraphics[width=0.45\textwidth]{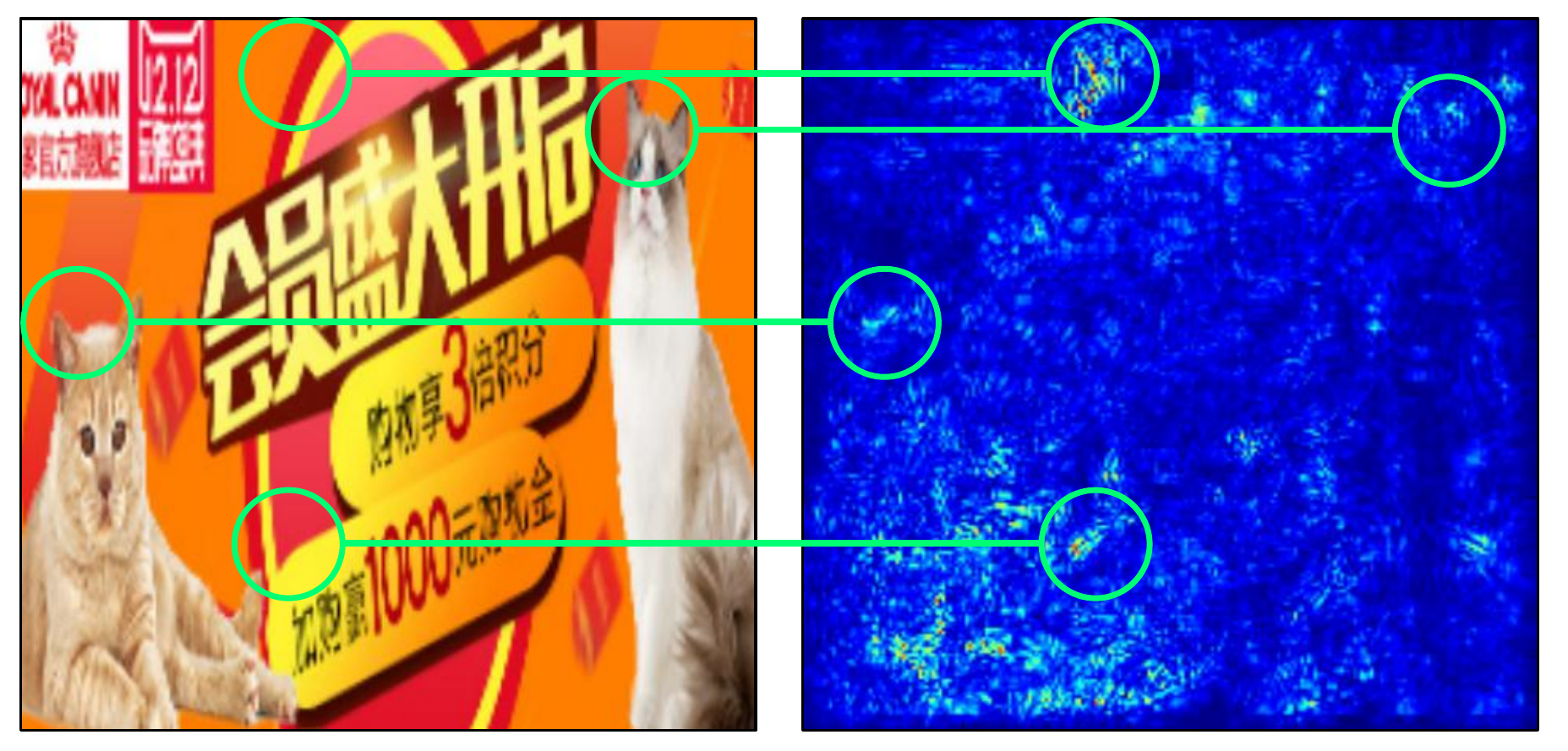}
	\caption{Saliency map of an image ad. We can see that cats, texture, and characters all have effect on the CTR. }
	\label{fig:vis_compare}
\end{figure}
\subsection{Visualizing the Convnet}
\begin{figure*}
	\centering
	\includegraphics[width=0.95\textwidth]{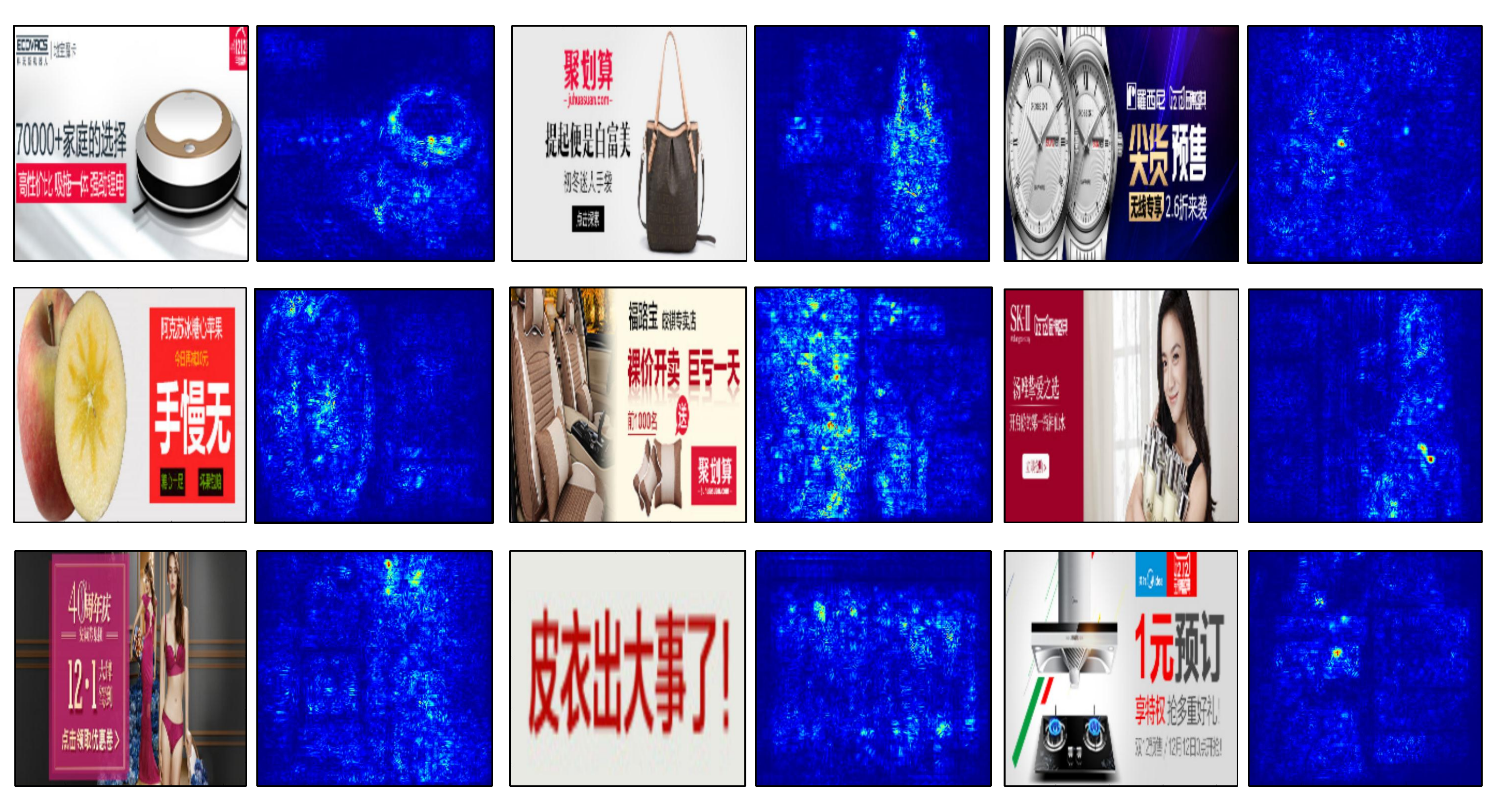}
	\caption{Saliency map of the image ads. Brighter area plays a more important role in effecting the CTR of ads.}
	\label{fig:vis}
\end{figure*}
Visualizing the CNN can help us better understand exactly what we have learned. In this section, we follow the saliency map visualization method used in \cite{simonyan2013deep}. We use a linear score model to approximate our DeepCTR for \emph{clicked} or \emph{not clicked}:
\begin{equation}\label{eq:approx}
	z(U) \approx w^{T}U + b, 
\end{equation}
where image $U$ is in the vectorized (one-dimension) form, and $w$ and $b$ are weight and bias of the model. Indeed, Eq \ref{eq:approx} can be seen as the first order Taylor expansion of our DeepCTR model. We use the magnitude
of elements of $w$ to show the importance of the corresponding pixels of $U$ for the \emph{clicked} probability. where $w$ is the derivative of $z$ with respect to the image $U$ at the point (image) $U_{0}$:
\begin{equation}
	w  = \frac{\partial z}{\partial U} \Big|_{U_{0}}
\end{equation}
In our case, the image $U$ is in RGB format and has three channels at pixel $U_{i,j}$. To derive a single class saliency value $M_{i,j}$ of each pixel, we take the maximum  absolute value of $w_{i,j}$ across RGB channels $c$: 
\begin{equation}
	M_{i,j} = \text{max}_{c} |w_{i,j}(c)|
\end{equation}

Some of the typical visualization examples are shown as heat maps  in Figure \ref{fig:vis}. In  these examples, brighter area plays a more important role in impacting the CTR of the ad. We can see main objects in ads are generally more important. However,  some low level features like texture, characters, and even background can have effect on the CTR of the ad. In another example (Figure \ref{fig:vis_compare}), it is more clearly to see that visual features in both high level and low level have effectiveness. From the visualization of the \emph{Convnet}, we can find that  the task of display ads CTR prediction is quite different from object classification where high level features dominate in the top layers.  It  also   gives an explanation why an end-to-end training can improve the model. Apparently, the \emph{Convnet} can be trained to extract features that are more particularly useful for CTR prediction.

The visualization provides us an intuitive understanding of the impact of visual features in ad images which  may be useful  for designers to make design choices. For example, we can decide whether add another model or not in an ad according to the saliency map of this model. 
\section{Conclusions}
CTR prediction plays an important role in online display advertising business. Accurate prediction of the CTR of ads not only increases the revenue of web publishers, also improves the user experience. In this paper we propose an end to end integrated deep network to predict the CTR of image ads.  It consists of \emph{Convnet}, \emph{Basicnet} and \emph{Combnet}. \emph{Convnet} is used to extract image features automatically while \emph{Basicnet} is used to reduce the dimensionality of basic features. \emph{Combnet} can learn complex and effective non-linear features from these two kinds of  features. The usage of sparse fully-connected layer and data sampling techniques speeds up the training process significantly.
We evaluate DeepCTR model on a 50 million real world dataset. The empirical result demonstrates 
the effectiveness and efficiency of our DeepCTR model. 
\vfill\eject
%
\bibliographystyle{abbrv}
\bibliography{sigproc}  
%
\end{document}